\documentclass[journal,twoside,web]{ieeecolor}

\usepackage{generic}

\usepackage{cite}
\usepackage{amsmath,amssymb,amsfonts}
\usepackage{graphicx}
\usepackage{algorithm}
\usepackage{algorithmic}
\usepackage{booktabs}
\usepackage{multirow}
\usepackage{url}

\providecommand{\refname}{References}


\makeatletter
\let\NAT@parse\undefined
\makeatother
\usepackage{hyperref}
\hypersetup{
    colorlinks=true,
    citecolor=blue,
    linkcolor=blue,
    urlcolor=blue
}

\graphicspath{{figures/}}

\usepackage{hyperref}

\hypersetup{
    colorlinks=true,
    citecolor=blue,
    linkcolor=blue,
    urlcolor=blue
}

\begin{document}

\title{GDBA-Net: A Gated Dual-Branch Adaptation of SAM2
for Robust Cross-Domain Nuclei Segmentation}

\author{
Biplab Poudel$^{1,2,3}$,
Betserai Albert Magarire$^{3}$,
Anish Kumar Shah$^{4}$,
Jiacheng Xie$^{1,2,5}$,
Dong Xu$^{1,2,5,*}$
\\[2pt]
{\small
$^{1}$Department of Electrical Engineering and Computer Science,
University of Missouri, Columbia, MO 65211, USA\\
$^{2}$Bond Life Sciences Center,
University of Missouri, Columbia, MO 65211, USA\\
$^{3}$Department of Applied Computing,
Lander University, Greenwood, SC 29649, USA\\
$^{4}$Department of Electronics and Computer Engineering,
Sagarmatha Engineering College, Tribhuvan University, Nepal\\
$^{5}$Health Informatics Institute, Morsani College of Medicine,
University of South Florida, Tampa, FL 33612, USA\\
$^{*}$Corresponding author. Email:
\texttt{dxu1@usf.edu}
}
}

\maketitle
\thispagestyle{empty}

% Abstract
\begin{abstract}

Accurate nuclei segmentation underpins quantitative computational pathology, yet remains challenging due to morphological variability, weak boundaries, nuclear crowding, and pronounced domain shifts arising from differences in tissue preparation, staining, and acquisition.  While deep learning and recent foundation models have improved segmentation performance under matched distributions with the original training data, their generalization across heterogeneous histopathology domains remains limited. We introduce \textbf{GDBA-Net}, a parameter-efficient nuclei segmentation framework built around the proposed Gated Dual-Branch Adapter (GDBA).  GDBA-Net adapts a frozen SAM2 encoder by integrating pretrained semantic representations with explicit frequency-derived structural priors. The complementary representations are independently transformed through Grid MLP branches and dynamically integrated using sample-adaptive channel-wise gating, selectively emphasizing structural cues relevant to nuclear delineation. Despite using only 4.3 million trainable parameters, GDBA-Net achieves state-of-the-art in-dataset performance and consistently superior cross-dataset generalization across six diverse nuclei segmentation benchmarks under fully supervised, cross-dataset, and five-shot adaptation protocols, outperforming competing CNN-, transformer-, and foundation model-based methods. Comprehensive ablations confirm that these gains arise from the coordinated design of semantic-frequency complementarity and adaptive gating rather than increased model capacity. By uniting transferable semantic representations with explicit structural priors, GDBA-Net provides a robust, annotation-efficient solution to domain variability in nuclei segmentation. \url{https://github.com/biplabpoudel25/GDBA_Net}

\end{abstract}


% Main manuscript
\section{Introduction}
\label{sec:introduction}

\IEEEPARstart{A}{ccurate} segmentation of cell nuclei from histopathological images constitutes a fundamental analytical step in computational pathology, serving as the gateway to downstream quantitative tasks, including cell counting, cell profiling, spatial transcriptomics mapping, and diagnostic decision support ~\cite{cheng2021hardboundary, mahbod2024generalization, coudray2018classification}. The proliferation of digital pathology and whole-slide imaging (WSI) in modern clinical workflows has intensified the demand for automated, robust nuclei segmentation pipelines that alleviate the substantial annotation burden on pathologists and enable high-throughput, reproducible morphological analysis \cite{evans2022wsi}. However, reliable nuclei delineation remains challenging because nuclei exhibit substantial variations in size, shape, appearance, and spatial organization, while weak boundaries, densely clustered cells, staining variability, and imaging artifacts further complicate segmentation \cite{graham2019hovernet}. More importantly, differences in tissue preparation, staining protocols, scanners, and acquisition conditions introduce pronounced domain shifts, causing models trained on one dataset to deteriorate when applied to unseen histopathological data.  

Traditional image processing methods, including watershed segmentation, adaptive thresholding, and handcrafted morphological descriptors, provided initial promise in controlled, homogeneous tissue settings \cite{irshad2014review} but were highly sensitive to staining variability, noise, and nuclear crowding, often requiring extensive parameter tuning and failing in complex tissue environments \cite{chowdhury2025federated}. Deep learning substantially advanced the field, beginning with U-Net \cite{ronneberger2015unet} and subsequently incorporating attention, multi-scale representations, instance-aware modeling, and transformer-based architectures \cite{graham2019hovernet, schlemper2019attention, zhou2020unetpp, schmidt2018stardist, chen2021transunet, cao2023swinunet}. Although these methods have achieved strong performance under matched training and testing distributions, robust generalization across heterogeneous datasets remains difficult. Their learned representations are largely driven by the source-domain appearance, making segmentation particularly vulnerable when staining characteristics, tissue morphology, or acquisition conditions differ from those encountered during training.

Vision foundation models provide a promising alternative by transferring representations learned from large and diverse image corpora. Segment Anything Model (SAM) \cite{kirillov2023sam} and its successor SAM2 \cite{ravi2024sam2} have demonstrated strong general-purpose segmentation capabilities, yet their direct transfer to fine-grained biomedical structures remains limited by the substantial gap between natural and histopathological images \cite{cui2024allinsam, vandeloo2025samcell}. Medical fine-tuning approaches like MedSAM \cite{ma2024medsam}, and MedSAM2 \cite{ma2025medsam2}, and parameter-efficient adaptation methods like SAMed \cite{zhang2023samed}, SAM-Adapter \cite{chen2023samadapter}, SAM2-Adapter \cite{chen2024sam2adapter}, and the Medical SAM Adapter \cite{wu2025medicalsamadapter}, have improved their applicability while reducing the need to train task-specific models from scratch. Nevertheless, efficient adaptation alone does not fully address the representational requirements of nuclei segmentation. 

High-level semantic features provide valuable contextual information, whereas accurate separation of densely packed nuclei also depends on fine structural cues associated with boundaries and local morphological transitions. Frequency-domain representations can emphasize such information \cite{li2023globalfrequency}, but high-frequency components may simultaneously amplify staining irregularities, texture variations, and acquisition noise. Thus, a central challenge is not simply introducing structural information, but determining how structural and semantic representations should be selectively balanced during adaptation. Existing adapter methods have explored task-specific information, including high-frequency components \cite{chen2024sam2adapter}, but the adaptive integration of separately transformed structural and semantic representations remains insufficiently explored to achieve robust nuclei segmentation across heterogeneous domains.

In this work, we present \textbf{GDBA-Net}, a SAM2-based framework for robust multi-domain nuclei segmentation. The framework is motivated by the complementary strengths of pretrained deep representations for semantic abstraction and frequency-derived representations for capturing fine structural information. Its central component is the proposed Gated Dual-Branch Adapter (GDBA), which uses branch-specific Grid MLP adapters to independently process semantic and frequency-derived structural features before integrating them through sample-adaptive channel-wise gating. This design selectively emphasizes structural cues relevant to nuclear delineation while preserving the pretrained knowledge of the frozen encoder. We evaluate GDBA-Net across six morphologically diverse benchmarks - Kumar \cite{kumar2017dataset}, MoNuSeg \cite{kumar2020monuseg}, CryoNuSeg \cite{mahbod2021cryonuseg}, CPM \cite{vu2019cpm}, TNBC \cite{naylor2019tnbc}, and CoNSeP \cite{graham2019hovernet},  under both in-dataset and rigorous cross-dataset generalization protocols.

The principal contributions of this paper are as follows:
\begin{enumerate}
    \item We introduce \textbf{GDBA-Net}, a parameter-efficient dual-branch 
    adaptation framework for SAM2 that jointly leverages pretrained semantic 
    representations and frequency-derived structural information for robust 
    nuclei segmentation.

    \item We develop a sample-adaptive channel-wise gating mechanism that dynamically balances semantic and structural information during adaptation. This design enables the model to selectively exploit fine structural cues while preserving transferable semantic knowledge, rather than relying on fixed or direct feature fusion.

    \item We demonstrate state-of-the-art performance and strong cross-domain generalization across six heterogeneous nuclei segmentation benchmarks under in-dataset, cross-dataset, and few-shot settings outperforming eleven competing CNN-, transformer-, and foundation model-based methods.

    \item We establish the effectiveness and efficiency of GDBA-Net through few-shot adaptation and systematic ablation studies. Additional experiments on retinal vessel and skin lesion segmentation further support the applicability of the proposed adaptation strategy beyond nuclei segmentation.
\end{enumerate}
\section{Related Work}
\label{sec:related_work}

\subsection{Deep Learning for Nuclei Segmentation}

Deep convolutional neural networks, anchored by the U-Net architecture \cite{ronneberger2015unet}, have become the dominant paradigm for biomedical image segmentation. In the nuclei segmentation literature, architectural innovations such as attention mechanisms \cite{oktay2018attentionunet}, dense skip connections \cite{isensee2021nnunet}, and multi-scale feature pyramids \cite{zhao2016pspnet} progressively improved performance on standard benchmarks. Task-specific frameworks addressing the challenge of overlapping nuclei further advanced the field: StarDist \cite{schmidt2018stardist} predicts object probabilities together with radial distances that define star-convex polygonal representation, while HoVer-Net \cite{graham2019hovernet} decomposes the segmentation task into horizontal and vertical gradient prediction branches to enable instance separation in crowded tissues. DAN-NucNet \cite{ahmad2023dannucnet} incorporates spatial and channel attention to enhance nuclear representations under heterogeneous histological conditions. Transformer-based architectures, including  \cite{chen2021transunet}, Swin-UNet \cite{cao2023swinunet}, HiFormer \cite{heidari2023hiformer}, and TransNuSeg \cite{he2023transnuseg}, subsequently improved the modeling of long-range dependencies and global-local context. Although these methods substantially improve representation learning and instance separation, these architectural advances do not inherently ensure robustness to domain shifts across tissue types, staining characteristics, preparation protocols, and acquisition conditions.

\subsection{Domain Shift and Generalization in Computational Pathology}

Domain shift remains a major challenge in computational pathology, where variations in staining, tissue preparation, scanners, and acquisition protocols can substantially alter image appearance and degrade performance on unseen data \cite{tellez2019stain, stacke2021domainshift}. Cross-domain evaluations have further demonstrated that strong performance within a single dataset does not necessarily translate across acquisition conditions or institutions \cite{aubreville2023midog}. Consequently, domain-generalization methods have explored appearance perturbation, representation learning, and frequency-based augmentation to reduce dependence on source-specific characteristics \cite{ouyang2023causality, li2023frequencymixed}.

\subsection{Parameter-Efficient Fine-Tuning of Foundational Models}

The emergence of segmentation foundation models has created new opportunities for transferring large-scale visual representations to medical imaging. SAM \cite{kirillov2023sam} introduced promptable general-purpose segmentation, while SAM2 \cite{ravi2024sam2} extended this paradigm using a hierarchical multi-scale encoder. However, large-scale pretraining on natural images does not eliminate the domain gap to specialized medical imagery, while full-model fine-tuning can impose substantial computational and memory requirements, particularly when task-specific annotations are limited. Parameter-efficient fine-tuning (PEFT) strategies, originally developed for natural language processing and later extended to vision models, have emerged as a principled solution. In medical imaging, approaches such as SAMed \cite{zhang2023samed} applied low-rank adaptation (LoRA) to the SAM image encoder, while SAM-Adapter \cite{chen2023samadapter}, SAM2-Adapter \cite{chen2024sam2adapter}, and the Medical SAM Adapter \cite{wu2025medicalsamadapter} introduce lightweight mechanisms for incorporating task- or domain-specific information while retaining much of the pretrained representation. MedSAM \cite{ma2024medsam} and MedSAM2 \cite{ma2025medsam2} instead rely on full or partial fine-tuning on curated medical datasets.

These approaches demonstrate that foundational models can be effectively specialized without developing task-specific architectures from scratch. However, they differ substantially in the number of trainable parameters, the form of task-specific information introduced, and the extent to which the pretrained backbone is modified. Moreover, most were developed for broad downstream or medical segmentation tasks rather than specifically addressing fine-grained nuclear structure and cross-dataset robustness in heterogeneous histopathology.

\subsection{Frequency-Domain Feature Integration}

Frequency-domain representations have been increasingly investigated to improve robustness to distribution shift. Fourier Domain Adaptation and related approaches have demonstrated that manipulating frequency components can improve domain adaptation and generalization, including in medical image segmentation. Frequency-aware architectures such as FRUNet \cite{zou2023frunet} have further shown the value of high-frequency information for capturing fine-grained structural details in biomedical segmentation. However, these approaches primarily employ frequency information for distribution manipulation, augmentation, or feature attention.
\section{Methodology}
\label{sec:methodology}

\subsection{Proposed Architecture Overview}
\label{subsec:architecture_overview}

\begin{figure*}
    \centering
    \includegraphics[width=0.95\textwidth]{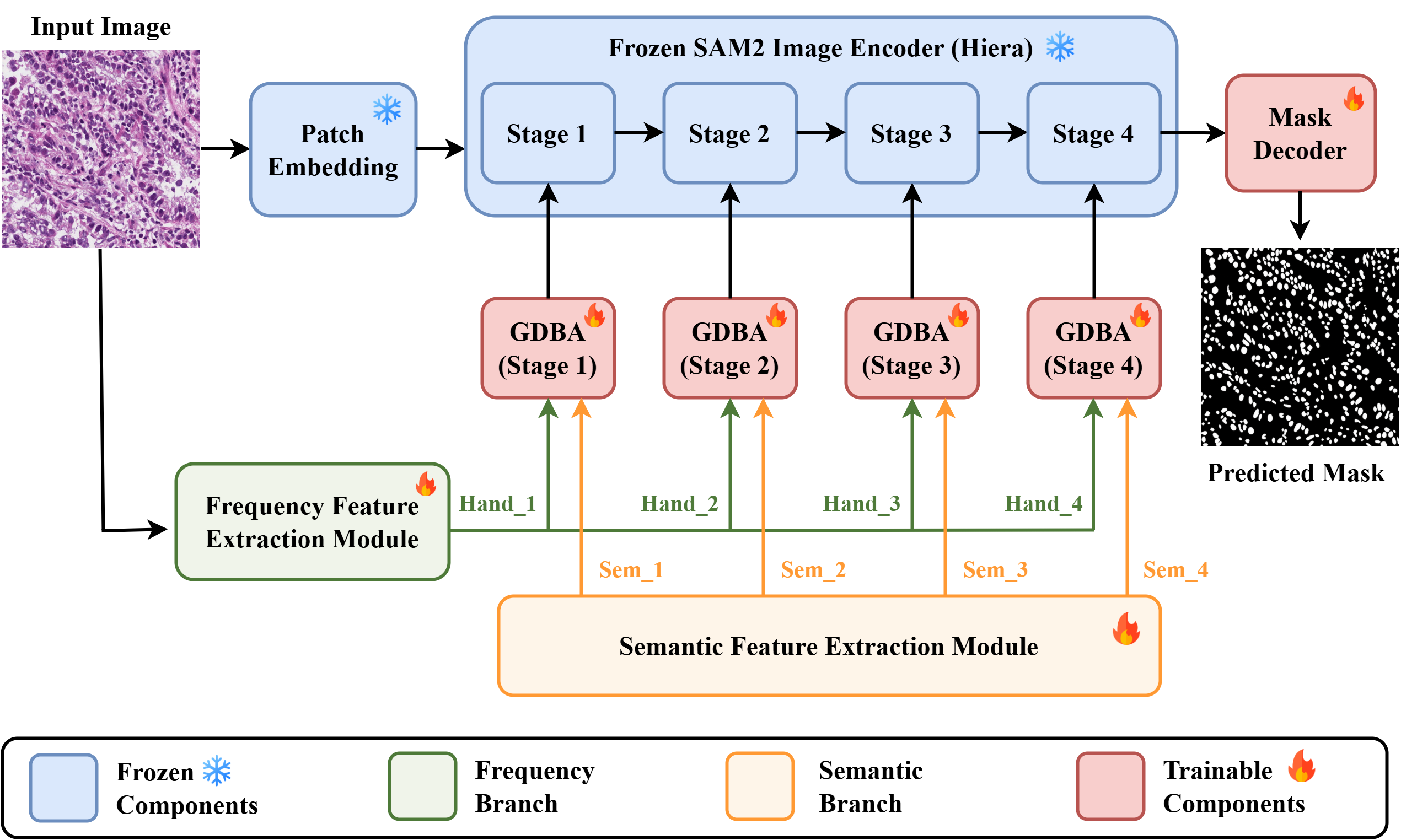}
    \caption{\textbf{Overall architecture of the proposed Gated Dual-Branch Adapter Network (GDBA-Net) framework for nuclei segmentation.} The input H\&E image is processed through multiple encoder stages, while the extracted frequency-derived structural and semantic features are integrated by the GDBA modules at the corresponding stages. The resulting representations are subsequently used to generate the predicted nuclei mask. The frequency and semantic feature extraction modules are detailed in Figure~\ref{fig:feature_extraction}.}
    \label{fig:architecture}
\end{figure*}

The proposed GDBA-Net framework, as shown in Figure~\ref{fig:architecture}, adapts the hierarchical image encoder of SAM2 for nuclei segmentation using the Gated Dual-Branch Adapter (GDBA). Given an input histopathology image $I \in \mathbb{R}^{B \times C_{in} \times H \times W}$ where $B$ is the batch size, $C_{in}=3$, and $H=W=1024$, the image is first converted into visual tokens using the SAM2 patch embeddings and positional embedding:
\begin{equation}
X_1 = PatchEmbed(I) + E_{pos}.
\end{equation}

The tokens are then progressively processed through the hierarchical encoder, where query pooling at stage transitions reduces the spatial resolution and increases the channel dimension, producing a hierarchical representation as:
\begin{equation}
X_s \in \mathbb{R}^{B \times H_s \times W_s \times C_s},
\qquad s \in \{1,2,3,4\}
\end{equation}
with $C_s \in \{144,288,576,1152\}$.

To adapt these pretrained representations for nuclei segmentation, GDBA integrates semantic information from the SAM2 encoder with frequency-derived structural information from the input image. The resulting adaptation features are injected into the corresponding SAM2 tokens through residual addition before the associated transformer blocks, enabling task-specific adaptation throughout the hierarchical encoder.

\begin{figure*}
    \centering
    \includegraphics[width=0.95\textwidth]{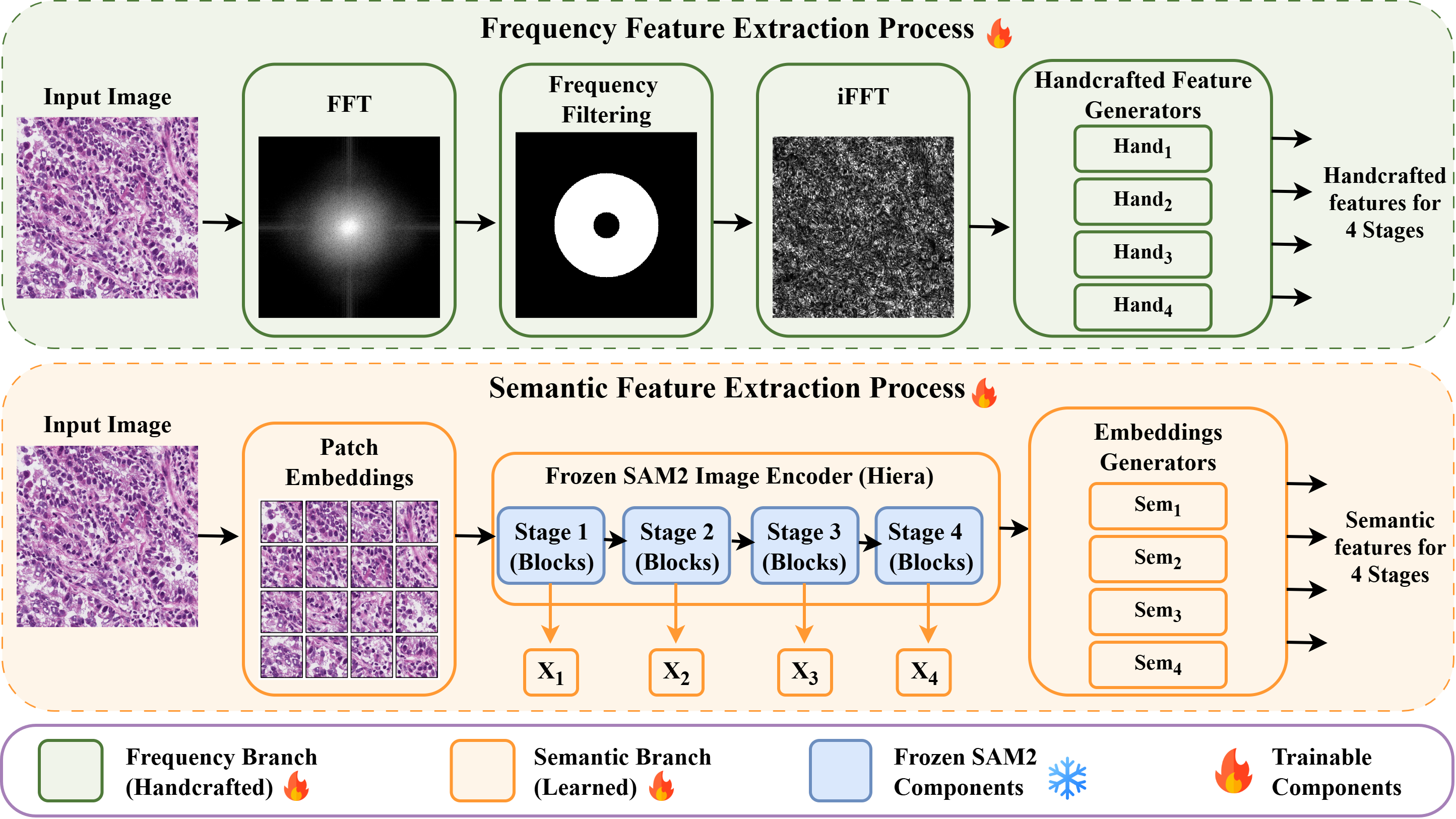}
    \caption{\textbf{Frequency and semantic feature extraction process used in GDBA framework.} The upper panel illustrates the frequency branch, whereas the lower panel shows the semantic feature extraction process. These complementary features are subsequently integrated by the GDBA modules.}
    \label{fig:feature_extraction}
\end{figure*}

\subsection{Semantic and Frequency-Derived Feature Construction}
\label{subsec:feature_construction}

The two complementary inputs to GDBA are constructed at multiple resolutions corresponding to the hierarchical feature levels of the SAM2 encoder, as described below.

\subsubsection{Semantic Representation}
\label{subsubsec:semantic_representation}

For each feature level $s$, the hierarchical SAM2 representation $X_s$ serves as the source of semantic integration. To obtain a parameter-efficient representation compatible with GDBA, $X_s$ is projected from the original encoder dimension $C_s$ into a lower-dimensional feature space:
\begin{equation}
Sem_s = X_s W_s^e + b_s^e,\qquad
Sem_s \in \mathbb{R}^{B \times H_s \times W_s \times C_s'}
\end{equation}
where $W_s^e$ and $b_s^e$ are learnable projection parameters. The resulting representation $Sem_s$ constitutes the semantic representation supplied to the semantic branch of GDBA.

\subsubsection{Frequency-Domain Handcrafted Feature Generation}
\label{subsubsec:frequency_features}

To explicitly capture robust structural cues often blurred by purely spatial learned features, we employ frequency-domain filtering via 2D Fast Fourier Transform:
\begin{equation}
\mathcal{F}(I)=\mathrm{FFTShift}(\mathrm{FFT2}(I)),\qquad
\mathcal{F}_h=\mathcal{F}(I)\odot(1-M)
\end{equation}
A binary mask $M$ suppresses the central 25\% of frequency coefficients to isolate high-frequency components. The filtered representation is transformed back to the spatial domain via inverse FFT:
\begin{equation}
I_h=\left|\mathrm{Re}\left[\mathrm{IFFT2}\left(\mathrm{IFFTShift}(\mathcal{F}_h)\right)\right]\right|
\end{equation}
Thus, low-frequency appearance information is suppressed while rapid spatial variations associated with boundaries, contours, and fine structural patterns are retained. Handcrafted features for each stage are then sequentially embedded:
\begin{equation}
Hand_s=PatchEmbed_s(Hand_{s-1})
\in\mathbb{R}^{B\times H_s\times W_s\times C_s'}
\end{equation}
where $C_s'=C_s/\alpha$, with $\alpha=32$ acts as a bottleneck scaling factor.

\subsection{Gated Dual Branch Adapter (GDBA)}
\label{subsec:gdba}

The GDBA module constitutes the core adaptation mechanism, inserted at each transformer block indexed by $(s,d)$, where $s$ denotes the stage and $d$ denotes the depth within the stage.

\subsubsection{Branch-specific Grid MLP Adapter (Branch Adapter)}
\label{subsubsec:grid_mlp}

At each transformer depth $d$, GDBA processes the semantic and frequency-derived representations through two independent \textbf{Grid MLP Adapters}. For an input $Z \in \mathbb{R}^{B \times H_s \times W_s \times C'_s}$, the Grid MLP Adapter is defined as:
\begin{equation}
\mathcal{A}(Z)=Z+\mathcal{D}_2
\left[W_2\,\mathcal{D}_1
\left(\mathrm{GELU}\left(W_1\mathrm{LN}(Z)+b_1\right)\right)+b_2\right]
\end{equation}

Here, $W_1$ expands the channel dimension from $C'_s$ to $4C'_s$, $W_2$ projects it back to $C'_s$, and $\mathcal{D}_1$ and $\mathcal{D}_2$ denote dropout operations. A residual connection preserves the original branch representation.

The MLP operates independently at every spatial grid location; hence, it performs nonlinear channel transformation without directly mixing neighboring spatial positions. Separate Grid MLP parameters are learned for the semantic and structural branches at each transformer depth:
\begin{equation}
Z_{s,d}^{\mathrm{sem}}=\mathcal{A}_{s,d}^{\mathrm{sem}}(Sem_s),
\qquad
Z_{s,d}^{\mathrm{freq}}=\mathcal{A}_{s,d}^{\mathrm{freq}}(Hand_s)
\end{equation}

The two adapted representations are concatenated along the channel dimension and reduced to the original adapter dimension using a depth-specific fusion layer:
\begin{equation}
F_{s,d}=\mathrm{GELU}
\left(
W_{s,d}^{f}
\left[
Z_{s,d}^{\mathrm{sem}}\Vert Z_{s,d}^{\mathrm{freq}}
\right]
+b_{s,d}^{f}
\right)
\end{equation}
where $F_{s,d}\in\mathbb{R}^{B\times H_s\times W_s\times C'_s}$ and $\Vert$ denotes channel-wise concatenation.

The fusion operation enables the model to learn optimal weighting between handcrafted structural cues and learned semantic representations, allowing it to dynamically emphasize whichever feature type is more informative for a given spatial location or image context.

\subsubsection{Adaptive Channel-Wise Gating}
\label{subsubsec:adaptive_gating}

Because frequency-derived information may contain useful nuclear boundaries as well as undesirable texture or acquisition artifacts, the fused representation is further regulated using an input-adaptive channel gate. Global average pooling is first performed over the spatial dimensions:
\begin{equation}
\bar{F}_{s,d}
=
\frac{1}{H_s W_s}
\sum_{h=1}^{H_s}
\sum_{w=1}^{W_s}
F_{s,d}(h,w)
\end{equation}

A depth-specific linear transformation followed by a sigmoid activation produces the channel gate:
\begin{equation}
g_{s,d}
=
\sigma\left(W_{s,d}^{g}\bar{F}_{s,d}+b_{s,d}^{g}\right),
\qquad
g_{s,d}\in[0,1]^{C'_s}
\end{equation}

The gate is then broadcast across spatial locations and applied to the fused representation:
\begin{equation}
\hat{F}_{s,d}(h,w,c)
=
F_{s,d}(h,w,c)\,g_{s,d}(c)
\end{equation}

Therefore, the implemented gate is sample-dependent and channel-wise, rather than spatially varying. Moreover, the gate operates on the fused representation rather than independently assigning weights to the semantic and frequency branches; cross-branch interaction is learned by the preceding fusion projection.

\begin{figure*}
    \centering
    \includegraphics[width=0.95\textwidth]{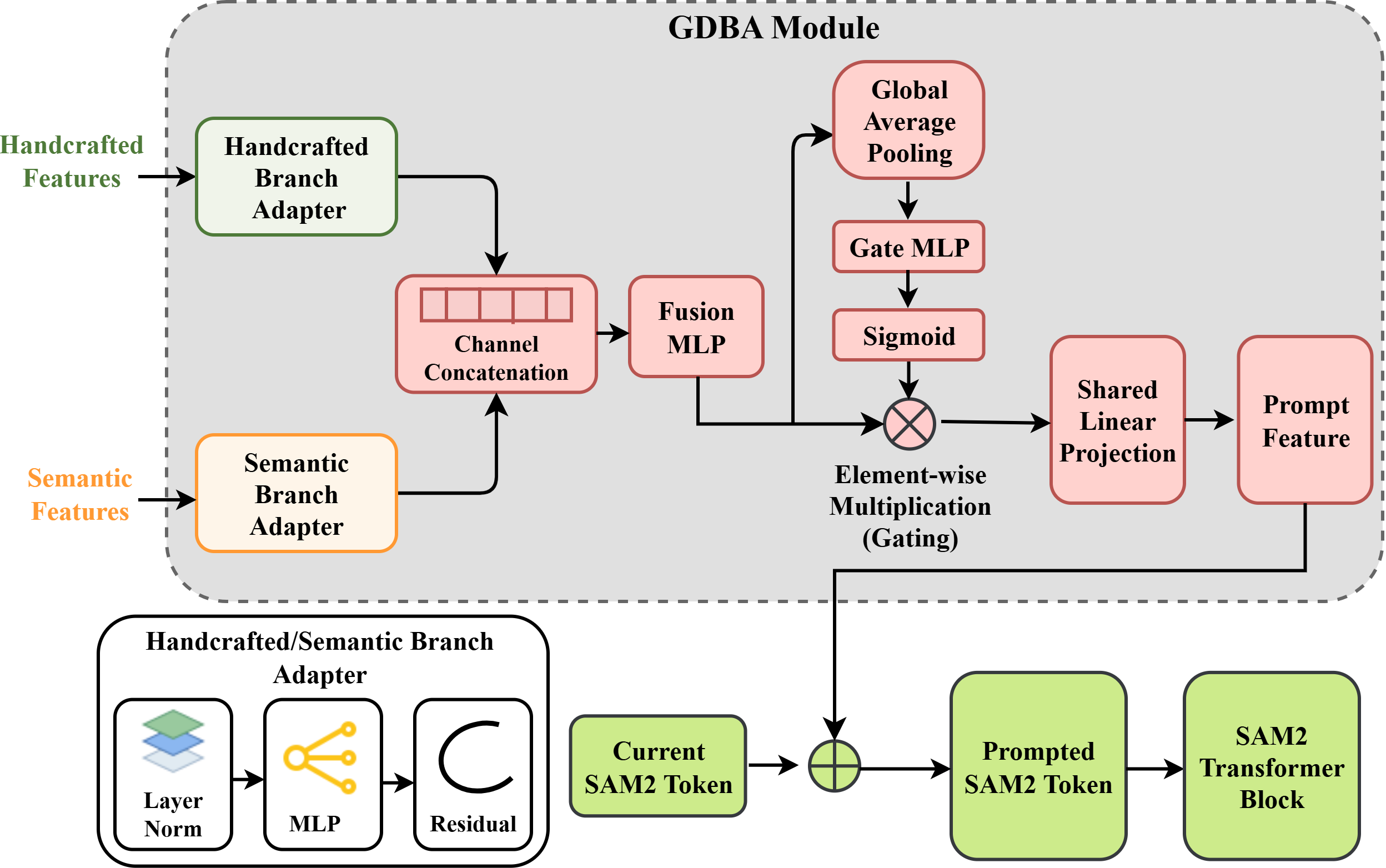}
    \caption{\textbf{Overview of the proposed Gated Dual-Branch Adapter (GDBA).} The upper light-gray shaded region illustrates the GDBA module, which integrates independently adapted handcrafted and semantic features to generate prompt features. The lower-left inset presents the branch adapter architecture, which is applied independently to the handcrafted and semantic feature streams. The generated prompts are combined with the current SAM2 tokens and passed to the corresponding transformer block.}
    \label{fig:gdba}
\end{figure*}

\subsubsection{Prompt Projection and Integration with SAM2}
\label{subsubsec:prompt_projection}

The gated representation remains in the compressed adapter space $C'_s$. A stage-specific linear projection maps it back to the native SAM2 embedding dimension:
\begin{equation}
P_{s,d}=W_s^p\hat{F}_{s,d}+b_s^p,
\qquad
P_{s,d}\in\mathbb{R}^{B\times H_s\times W_s\times C_s}
\end{equation}

Unlike the branch adapters, fusion layers, and gating layers, which are depth-specific, the final projection $W_s^p$ is shared among the GDBA modules operating at the same feature resolution.

The generated feature is then injected through residual addition into the current SAM2 tokens immediately before the corresponding Hiera transformer block:
\begin{equation}
\tilde{X}_{s,d}=X_{s,d}+P_{s,d}
\end{equation}
followed by:
\begin{equation}
X_{s,d+1}=T_{s,d}(\tilde{X}_{s,d})
\end{equation}
where $T_{s,d}$ represents the corresponding pretrained Hiera block. Consequently, GDBA does not replace the SAM2 representation; instead, it learns a task-specific residual modification that introduces complementary semantic and structural information into the existing feature stream.

This process repeats for each depth $d$ within stage $s$, with the GDBA module progressively refining the feature representations as they propagate through the encoder hierarchy. The final segmentation mask is generated by the SAM2 Mask Decoder.

\subsection{Training Objective}
\label{subsec:training_objective}

Nuclei typically occupy a substantially smaller region than the histopathological background, resulting in considerable foreground-background imbalance. The model is therefore optimized using the Tversky loss. Given true positives TP, false positives FP, and false negatives FN, the Tversky index is defined as:
\begin{equation}
TI=\frac{TP}{TP+\alpha_T FP+\beta_T FN},
\qquad
\mathcal{L}_{\mathrm{Tversky}}=1-TI
\end{equation}
where $\alpha_T$ and $\beta_T$ control the relative penalties assigned to false-positive and false-negative predictions. The optimization objective is $\mathcal{L}_{\mathrm{Tversky}}=1-TI$.

During adaptation, the pretrained SAM2 image-encoder parameters are kept fixed, while the GDBA and decoder-related trainable parameters are optimized. Training is performed using the Adam optimizer with a learning rate of $1\times10^{-4}$ for 200 epochs.

\section{EXPERIMENTAL SETUP}
\label{sec:experimental_setup}

\subsection{Datasets}
\label{subsec:datasets}

We evaluate the proposed framework on six publicly available H\&E-stained nuclei segmentation benchmarks: Kumar~\cite{kumar2017dataset}, MoNuSeg~\cite{kumar2020monuseg}, CryoNuSeg~\cite{mahbod2021cryonuseg}, Combined CPM~\cite{vu2019cpm}, TNBC~\cite{naylor2019tnbc}, and CoNSeP~\cite{graham2019hovernet}. These datasets encompass diverse tissue types, image resolutions, and preparation protocols, providing a comprehensive setting for evaluating both in-dataset performance and cross-dataset generalization. Table~\ref{tab:datasets} summarizes the characteristics of each dataset along with the training, validation, and testing partitions used in our experiments.

The validation sets were separated from the available training data and used for model selection, while the test sets were reserved exclusively for final evaluation. For the Combined CPM dataset, CPM-15 and CPM-17 were merged by combining their respective training, validation, and test partitions. The same fixed partitions were used across all competing methods to ensure consistent comparison.

\begin{table}[!t]
\centering
\caption{Summary of benchmark datasets used for nuclei segmentation evaluation.}
\label{tab:datasets}

\scriptsize
\setlength{\tabcolsep}{2.2pt}
\renewcommand{\arraystretch}{1.05}

\begin{tabular}{@{}l p{1.45cm} c c c c@{}}
\hline
\textbf{Dataset} &
\textbf{Tissue / Condition} &
\textbf{Images} &
\textbf{Resolution} &
\textbf{Nuclei} &
\textbf{\begin{tabular}[c]{@{}c@{}}Split\\(Train/Val/Test)\end{tabular}} \\
\hline

Kumar~\cite{kumar2017dataset} &
Multi-organ (7 organs) &
30 &
1000$\times$1000 &
21,623 &
12/4/14 \\

MoNuSeg~\cite{kumar2020monuseg} &
Multi-organ (9 organs) &
51 &
1000$\times$1000 &
$\sim$28,500 &
33/7/11 \\

CryoNuSeg~\cite{mahbod2021cryonuseg} &
Multi-organ (10 organs) &
30 &
512$\times$512 &
$\sim$7,596 &
15/5/10 \\

Combined CPM~\cite{vu2019cpm} &
Brain, Head \& Neck, Lung &
79 &
Variable &
$\sim$10,000 &
30/10/39 \\

TNBC~\cite{naylor2019tnbc} &
Breast (TNBC subtype) &
50 &
512$\times$512 &
4,022 &
32/8/10 \\

CoNSeP~\cite{graham2019hovernet} &
Colorectal adenocarcinoma &
41 &
1000$\times$1000 &
24,319 &
22/5/14 \\

\hline
\end{tabular}
\end{table}

\subsection{Experimental Protocols}
\label{subsec:experimental_protocols}

We evaluate GDBA-Net under two complementary settings: in-dataset evaluation and cross-dataset generalization. In the in-dataset setting, models are trained and evaluated using the training and testing partitions of the same dataset. In the cross-dataset setting, models trained on a source dataset are directly evaluated on the unseen target dataset without target-domain fine-tuning, providing a measure of robustness under domain shift. To further examine the effect of limited target-domain supervision, we compare direct cross-dataset transfer with 5-shot target-domain adaptation and full target-domain training. In the 5-shot setting, five target-domain images are used for training, a small subset is retained for validation, and the remaining images are used for testing. Additionally, the applicability of GDBA-Net beyond nuclei segmentation is evaluated on distinct biomedical segmentation tasks.

\subsection{Evaluation Metrics}
\label{subsec:evaluation_metrics}

Performance is quantified using Dice Similarity Coefficient (DSC), Jaccard Index (IoU), Precision, and Specificity. Using the pixel-level TP, FP, and FN quantities defined above, and denoting the number of true-negative background pixels by TN, the metrics are computed as:

\begin{equation}
\begin{aligned}
\mathrm{DSC} &= \frac{2TP}{2TP+FP+FN}, &
\mathrm{IoU} &= \frac{TP}{TP+FP+FN},\\[-2pt]
\mathrm{Prec.} &= \frac{TP}{TP+FP}, &
\mathrm{Spec.} &= \frac{TN}{TN+FP}.
\end{aligned}
\end{equation}

Dice and Jaccard quantify the spatial overlap between the predicted and ground-truth nuclei masks. Precision reflects false-positive foreground predictions, whereas Specificity measures the model's ability to correctly identify background pixels. Together, these metrics provide complementary measures of segmentation overlap and pixel-level classification performance.
\section{RESULTS}
\label{sec:results}

We evaluate GDBA-Net across heterogeneous nuclei segmentation benchmarks to examine its segmentation accuracy, robustness to domain shift, and ability to adapt under limited target-domain supervision. Comparisons with conventional CNN-, transformer-, nuclei-specific, and foundation model-based approaches are used to assess performance across varying tissue types, preparation protocols, and imaging characteristics.

\subsection{In-dataset evaluation}
\label{subsec:indataset_evaluation}

Table~\ref{tab:indataset} presents the in-dataset performance on MoNuSeg, Kumar, and CryoNuSeg, where models are trained and tested on the same dataset. GDBA-Net achieves the highest Dice and Jaccard scores on all three datasets, demonstrating consistently accurate segmentation across diverse tissue types, image characteristics, and preparation protocols. The comparison further shows that architectural complexity or foundation-model pretraining alone does not guarantee strong segmentation performance. While nnU-Net, HoVer-Net, and DanNucNet provide competitive results, transformer-based models and directly adapted SAM variants show less consistent performance across datasets. In contrast, GDBA-Net maintains strong performance under both FFPE and cryo-sectioned imaging conditions, supporting the effectiveness of combining SAM2 semantic representations with frequency-derived structural information. Notably, the 5-shot GDBA-Net achieves a 0.8359 Dice on CryoNuSeg, surpassing fully trained foundational models like MedSAM. Overall, these results establish the proposed method as a robust and annotation-efficient approach for in-dataset nuclei segmentation.

\begin{table*}[!t]
\centering
\caption{In-dataset nuclei segmentation performance comparison of GDBA-Net and competing methods on the MoNuSeg, CryoNuSeg, and Kumar datasets. Best and second-best results are highlighted in bold and underline, respectively.}
\label{tab:indataset}

\scriptsize
\setlength{\tabcolsep}{2.7pt}
\renewcommand{\arraystretch}{1.05}

\begin{tabular}{lcccccccccccc}
\hline
\multirow{2}{*}{\textbf{Models}} &
\multicolumn{4}{c}{\textbf{MoNuSeg}} &
\multicolumn{4}{c}{\textbf{Kumar}} &
\multicolumn{4}{c}{\textbf{CryoNuSeg}} \\
\cline{2-13}
&
\textbf{Dice} & \textbf{Jaccard} & \textbf{Precision} & \textbf{Specificity} &
\textbf{Dice} & \textbf{Jaccard} & \textbf{Precision} & \textbf{Specificity} &
\textbf{Dice} & \textbf{Jaccard} & \textbf{Precision} & \textbf{Specificity} \\
% \textbf{Dice} & \textbf{Jaccard} & \textbf{Prec.} & \textbf{Spec.} \\
\hline

U-Net~\cite{ronneberger2015unet}
& 0.8021 & 0.6712 & 0.7929 & 0.9411
& 0.7278 & 0.5775 & 0.6332 & 0.8149
& 0.7737 & 0.6349 & 0.7400 & 0.8531 \\

Attn. U-Net~\cite{oktay2018attentionunet}
& 0.7903 & 0.6551 & 0.7615 & 0.9264
& 0.7609 & 0.6212 & 0.7451 & 0.8833
& 0.7596 & 0.6176 & 0.7164 & 0.8410 \\

nnU-Net~\cite{isensee2021nnunet}
& 0.8218 & 0.7091 & 0.8015 & 0.9396
& \underline{0.8199} & \underline{0.6966} & 0.8177 & \underline{0.9424}
& 0.8108 & 0.6845 & 0.8144 & 0.9186 \\

PSPNet~\cite{zhao2016pspnet}
& 0.5922 & 0.4311 & 0.4347 & 0.6061
& 0.6014 & 0.4348 & 0.4402 & 0.6274
& 0.6805 & 0.5204 & 0.5361 & 0.6042 \\

HoVer-Net~\cite{graham2019hovernet}
& 0.8208 & 0.7086 & 0.7996 & 0.9399
& 0.7981 & 0.6749 & 0.7741 & 0.9223
& 0.8284 & 0.7125 & 0.8452 & 0.9231 \\

NucleiSegNet~\cite{lal2021nucleisegnet}
& 0.7681 & 0.6251 & 0.7195 & 0.9095
& 0.7701 & 0.6283 & 0.7541 & 0.9221
& 0.8023 & 0.6756 & 0.8478 & 0.9282 \\

DanNucNet~\cite{ahmad2023dannucnet}
& \underline{0.8288} & \underline{0.7101} & \underline{0.8244} & 0.9452
& \underline{0.8199} & 0.6953 & 0.8169 & 0.9327
& 0.8033 & 0.6755 & 0.7904 & 0.9027 \\

TransNuSeg~\cite{he2023transnuseg}
& 0.7682 & 0.6257 & 0.7295 & 0.9145
& 0.7556 & 0.5975 & 0.7050 & 0.8690
& 0.7745 & 0.6355 & 0.7305 & 0.8563 \\

HiFormer~\cite{heidari2023hiformer}
& 0.6703 & 0.5060 & 0.7308 & 0.9322
& 0.6340 & 0.4669 & 0.6410 & 0.8984
& 0.7371 & 0.5843 & 0.8056 & 0.9226 \\

SAM~\cite{kirillov2023sam}
& 0.7842 & 0.6470 & 0.7991 & \underline{0.9462}
& 0.7091 & 0.5627 & 0.6838 & 0.8990
& 0.7416 & 0.5949 & 0.7571 & 0.9003 \\

MedSAM~\cite{ma2024medsam}
& 0.7811 & 0.6420 & 0.7552 & 0.9250
& 0.7617 & 0.6170 & 0.7746 & 0.7620
& 0.8092 & 0.6814 & 0.8063 & 0.9104 \\

GDBA-Net$_{\text{5-shot}}$
& 0.8262 & 0.7050 & 0.8220 & 0.9414
& 0.8180 & 0.6942 & \textbf{0.8390} & \textbf{0.9494}
& \underline{0.8359} & \underline{0.7205} & \textbf{0.8725} & \textbf{0.9436} \\

GDBA-Net
& \textbf{0.8408} & \textbf{0.7259} & \textbf{0.8388} & \textbf{0.9516}
& \textbf{0.8201} & \textbf{0.6972} & \underline{0.8104} & 0.9365
& \textbf{0.8481} & \textbf{0.7381} & \underline{0.8519} & \underline{0.9286} \\

\hline
\end{tabular}
\end{table*}

\subsection{Cross-dataset generalization}
\label{subsec:crossdataset_generalization}

Table~\ref{tab:crossdataset} presents cross-dataset generalization by training on a single source dataset and directly testing it on unseen target datasets without target-domain fine-tuning. Across the eight source-target combinations, GDBA-Net achieves the highest Dice and Jaccard scores in seven settings and ranks second in the remaining CoNSeP-to-CPM evaluation. This consistent performance across multiple source-target combinations indicates robustness to variations in tissue composition, staining appearance, image resolution, and preparation protocol.

\begin{table*}[!t]
\centering
\caption{Cross-dataset generalization performance of GDBA-Net and competing methods trained on MoNuSeg, CryoNuSeg, or CoNSeP and directly evaluated on the unseen Combined CPM, TNBC, and Combined CoNSeP test datasets. For each source--target dataset pair and metric, the best and second-best results are highlighted in bold and underline, respectively.}
\label{tab:crossdataset}

\scriptsize
\setlength{\tabcolsep}{2.1pt}
\renewcommand{\arraystretch}{1.05}

\begin{tabular}{c|l|cccc|cccc|cccc}
\hline
&
\multirow{2}{*}{\textbf{Methods}} &
\multicolumn{12}{c}{\textbf{Test dataset}} \\
\cline{3-14}
&
&
\multicolumn{4}{c|}{\textbf{Combined CPM}} &
\multicolumn{4}{c|}{\textbf{TNBC}} &
\multicolumn{4}{c}{\textbf{Combined CoNSeP}} \\
\cline{3-14}
&
&
\textbf{Dice} & \textbf{Jaccard} & \textbf{Precision} & \textbf{Specificity} &
\textbf{Dice} & \textbf{Jaccard} & \textbf{Precision} & \textbf{Specificity} &
\textbf{Dice} & \textbf{Jaccard} & \textbf{Precision} & \textbf{Specificity} \\
\hline

% ==================== Trained on MoNuSeg ====================
\multirow{12}{*}{\rotatebox[origin=c]{90}{\textbf{Trained on MoNuSeg}}}
& U-Net
& 0.7329 & 0.5878 & 0.8281 & 0.9640
& 0.6732 & 0.5187 & 0.7145 & 0.9642
& 0.6236 & 0.4678 & 0.5698 & 0.9000 \\

& Attn U-Net
& 0.7262 & 0.5799 & 0.8170 & 0.9589
& \underline{0.7027} & 0.5529 & 0.7438 & 0.9629
& 0.6003 & 0.4477 & 0.5147 & 0.8568 \\

& nnU-Net
& 0.7801 & 0.6604 & 0.8004 & 0.9440
& 0.7025 & \underline{0.5712} & 0.6132 & 0.9065
& \underline{0.6322} & 0.4854 & 0.5444 & 0.8407 \\

& PSPNet
& 0.6062 & 0.4565 & 0.4674 & 0.6863
& 0.2606 & 0.1634 & 0.1639 & 0.1928
& 0.4142 & 0.2844 & 0.2900 & 0.4596 \\

& HiFormer
& 0.7386 & 0.5886 & 0.7955 & 0.9600
& 0.6435 & 0.4870 & 0.6245 & 0.9562
& 0.5244 & 0.3646 & 0.5363 & 0.9203 \\

& HoVer-Net
& 0.8115 & 0.6949 & 0.8663 & 0.9659
& 0.6508 & 0.5370 & 0.7810 & \underline{0.9777}
& 0.6191 & \underline{0.4966} & 0.5971 & 0.9072 \\

& NucleiSegNet
& 0.7686 & 0.6474 & 0.8088 & 0.9485
& 0.6773 & 0.5474 & 0.5965 & 0.8955
& 0.6089 & 0.4600 & 0.5326 & 0.8400 \\

& DanNucNet
& \underline{0.8399} & \underline{0.7282} & \textbf{0.9165} & \textbf{0.9835}
& 0.5338 & 0.4039 & \textbf{0.8824} & \textbf{0.9831}
& 0.6101 & 0.4709 & \textbf{0.7424} & 0.8894 \\

& TransNuSeg
& 0.7466 & 0.6016 & 0.7967 & 0.9570
& 0.6578 & 0.5025 & 0.6241 & 0.9486
& 0.6175 & 0.4639 & 0.5572 & 0.8954 \\

& SAM
& 0.7094 & 0.5615 & 0.7248 & 0.9350
& 0.6409 & 0.5012 & 0.6190 & 0.9514
& 0.4995 & 0.3548 & 0.4926 & 0.8937 \\

& MedSAM
& 0.7649 & 0.6242 & 0.8260 & 0.9662
& 0.6261 & 0.4840 & 0.5803 & 0.9394
& 0.6124 & 0.4673 & 0.6065 & \underline{0.9240} \\

& GDBA
& \textbf{0.8558} & \textbf{0.7504} & \underline{0.8922} & \underline{0.9741}
& \textbf{0.8146} & \textbf{0.6936} & \underline{0.8127} & 0.9715
& \textbf{0.6833} & \textbf{0.5351} & \underline{0.7253} & \textbf{0.9565} \\
\hline

% ==================== Trained on CryoNuSeg ====================
\multirow{12}{*}{\rotatebox[origin=c]{90}{\textbf{Trained on CryoNuSeg}}}
& U-Net
& 0.6536 & 0.5177 & 0.8863 & 0.9752
& 0.2179 & 0.1311 & 0.8927 & \underline{0.9975}
& 0.1966 & 0.1184 & \underline{0.7858} & \underline{0.9898} \\

& Attn U-Net
& 0.6593 & 0.5131 & 0.8275 & 0.9607
& 0.3279 & 0.2093 & 0.8130 & 0.9938
& 0.3319 & 0.2124 & 0.7623 & 0.9841 \\

& nnU-Net
& 0.7819 & 0.6516 & 0.7476 & 0.9330
& 0.5602 & 0.4221 & 0.4599 & 0.8704
& 0.5603 & 0.4143 & 0.4916 & 0.8563 \\

& PSPNet
& 0.5678 & 0.4122 & 0.4630 & 0.7653
& 0.2926 & 0.1818 & 0.3527 & 0.9370
& 0.2727 & 0.1650 & 0.3194 & 0.8681 \\

& HiFormer
& 0.6147 & 0.4626 & 0.9001 & 0.9864
& 0.4292 & 0.2905 & 0.8616 & 0.9934
& 0.1520 & 0.0861 & 0.6602 & 0.9871 \\

& HoVer-Net
& 0.7227 & 0.5968 & 0.9004 & 0.9792
& 0.3344 & 0.2209 & \underline{0.9155} & 0.9952
& 0.5561 & 0.3977 & 0.7160 & 0.9628 \\

& NucleiSegNet
& 0.6326 & 0.5133 & \underline{0.9311} & \textbf{0.9898}
& 0.5089 & 0.3749 & 0.8929 & 0.9917
& \underline{0.6331} & \underline{0.4823} & 0.6839 & 0.9154 \\

& DanNucNet
& 0.6256 & 0.4946 & \textbf{0.9345} & \underline{0.9883}
& 0.0928 & 0.0516 & \textbf{0.9375} & \textbf{0.9994}
& 0.1980 & 0.1183 & \textbf{0.8152} & \textbf{0.9915} \\

& TransNuSeg
& 0.7114 & 0.5684 & 0.7804 & 0.7114
& 0.4344 & 0.2968 & 0.8742 & 0.4344
& 0.5354 & 0.3759 & 0.666 & 0.5354 \\

& SAM
& 0.7283 & 0.5830 & 0.7397 & 0.7283
& \underline{0.6520} & \underline{0.5065} & 0.6199 & 0.6520
& 0.4921 & 0.3438 & 0.4600 & 0.4921 \\

& MedSAM
& \underline{0.7932} & \underline{0.6611} & 0.8150 & 0.9584
& 0.6370 & 0.4863 & 0.5945 & 0.9439
& 0.5026 & 0.3582 & 0.4801 & 0.8632 \\

& GDBA
& \textbf{0.8266} & \textbf{0.7126} & 0.8744 & 0.9706
& \textbf{0.8045} & \textbf{0.6829} & 0.8271 & 0.9723
& \textbf{0.6557} & \textbf{0.5051} & 0.6649 & 0.9368 \\
\hline

% ==================== Trained on CoNSeP ====================
\multirow{12}{*}{\rotatebox[origin=c]{90}{\textbf{Trained on CoNSeP}}}
& U-Net
& 0.3961 & 0.2722 & 0.8602 & 0.9800
& 0.5443 & 0.4008 & 0.7971 & \textbf{0.9926}
& \multicolumn{4}{c}{\multirow{12}{*}{-----}} \\

& Attn U-Net
& 0.2888 & 0.1977 & 0.7817 & 0.9704
& 0.6611 & 0.5100 & 0.8225 & 0.9705
& \multicolumn{4}{c}{} \\

& nnU-Net
& 0.7912 & 0.6594 & 0.8832 & 0.9802
& \underline{0.7823} & \underline{0.6512} & 0.7929 & 0.9689
& \multicolumn{4}{c}{} \\

& PSPNet
& 0.4179 & 0.2883 & 0.3428 & 0.7109
& 0.4542 & 0.3123 & 0.3181 & 0.7307
& \multicolumn{4}{c}{} \\

& HiFormer
& 0.4441 & 0.3067 & 0.9078 & 0.9840
& 0.6274 & 0.4758 & 0.7736 & 0.9744
& \multicolumn{4}{c}{} \\

& HoVer-Net
& \textbf{0.8187} & \textbf{0.6979} & \underline{0.9293} & \underline{0.9845}
& 0.6055 & 0.4575 & \textbf{0.9458} & \underline{0.9916}
& \multicolumn{4}{c}{} \\

& NucleiSegNet
& 0.6509 & 0.5153 & 0.9228 & 0.9781
& 0.7207 & 0.5849 & 0.8695 & 0.9733
& \multicolumn{4}{c}{} \\

& DanNucNet
& 0.5370 & 0.4056 & 0.8913 & 0.9767
& 0.7047 & 0.5604 & 0.8433 & 0.9663
& \multicolumn{4}{c}{} \\

& TransNuSeg
& 0.6367 & 0.5038 & 0.8890 & 0.9715
& 0.5766 & 0.4311 & 0.8290 & 0.9824
& \multicolumn{4}{c}{} \\

& SAM
& 0.5984 & 0.4367 & 0.5957 & 0.8920
& 0.6519 & 0.4990 & 0.6726 & 0.9549
& \multicolumn{4}{c}{} \\

& MedSAM
& 0.7315 & 0.5797 & 0.8376 & 0.9671
& 0.6936 & 0.5374 & 0.7255 & 0.9659
& \multicolumn{4}{c}{} \\

& GDBA
& \underline{0.7936} & \underline{0.6624} & \textbf{0.9361} & \textbf{0.9858}
& \textbf{0.7853} & \textbf{0.6521} & \underline{0.8729} & 0.9785
& \multicolumn{4}{c}{} \\

\hline
\end{tabular}
\end{table*}

When trained on MoNuSeg, GDBA-Net achieves the best overlap performance on all three unseen datasets, with particularly clear improvements on TNBC (Dice: 0.8146), where methods such as nnU-Net and U-Net obtained substantially lower performance. Although DanNucNet remains competitive on CPM, its Dice decreases to 0.5338 on TNBC, demonstrating that strong transfer to one target dataset does not necessarily ensure consistent cross-domain generalization. A similar pattern is observed when CryoNuSeg is used as the source domain, with GDBA-Net achieving the highest Dice and Jaccard scores across the evaluated target datasets. Several competing models retain high Precision or Specificity under this shift but exhibit considerably lower overlap scores, suggesting conservative predictions that fail to recover substantial portions of the nuclear regions. In contrast, GDBA-Net maintains a better balance between nuclear detection and background discrimination.

A more challenging pattern emerges when CoNSeP is used as the source domain. Although HoVer-Net achieves the highest overlap on CPM, GDBA-Net produces the highest precision and specificity, indicating fewer false-positive predictions. More importantly, GDBA-Net regains the leading overlap performance on the CoNSeP to TNBC transfer, whereas HoVer-Net shows a substantial decrease despite its strong performance on CPM. This contrast highlights that success on a single dataset doesn’t necessarily indicate robust cross-domain generalization.

\begin{figure*}[!t]
    \centering
    \includegraphics[width=0.95\textwidth]{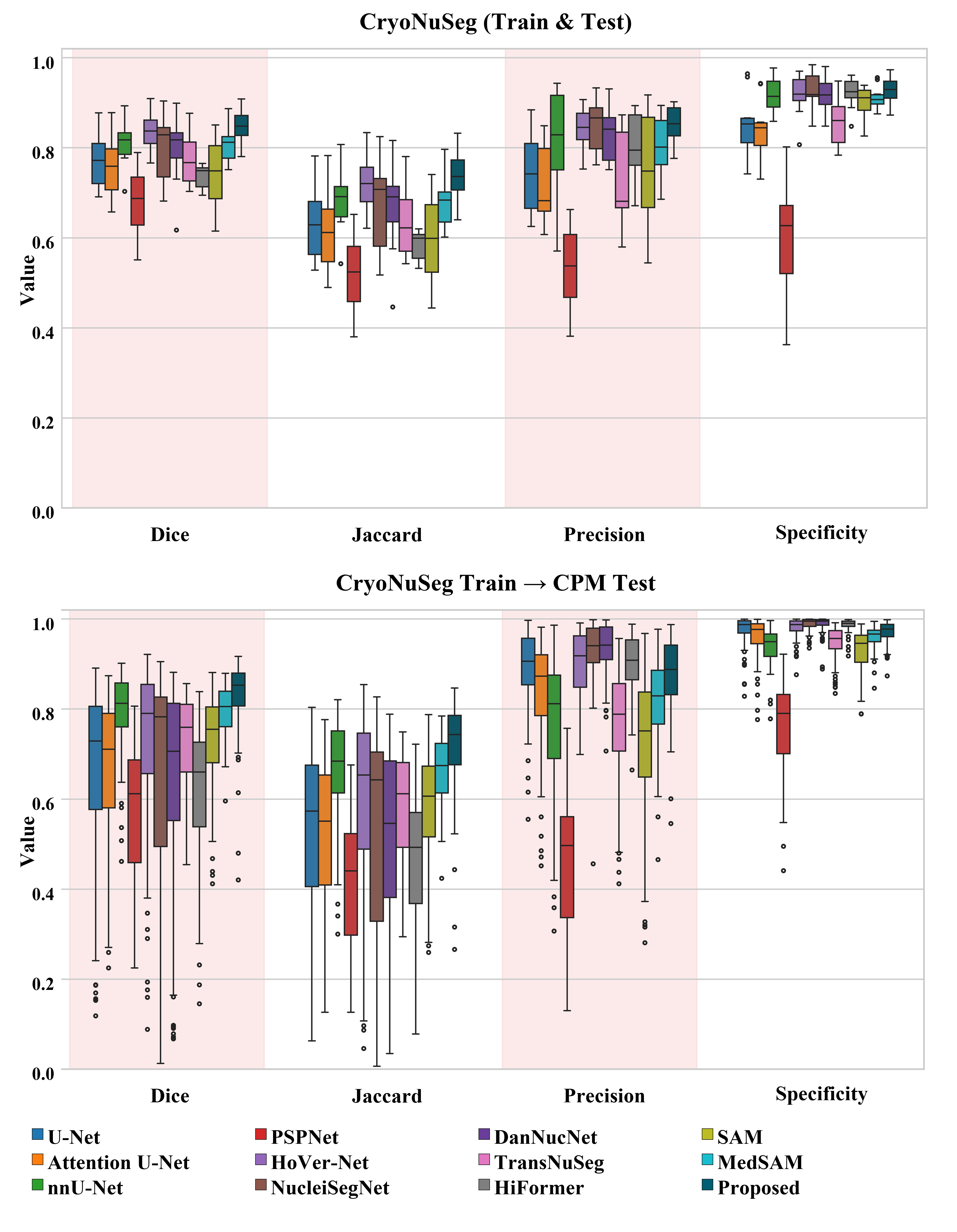}
    \caption{\textbf{Boxplot-based comparison of nuclei segmentation performance under in-dataset and cross-dataset evaluation.} The top panel presents results when models are trained and tested on CryoNuSeg, whereas the bottom panel shows cross-dataset performance when models are trained on CryoNuSeg and evaluated on CPM dataset. Boxes represent the interquartile range, horizontal lines indicate the median, whiskers show the non-outlier range, and circular markers denote outlier samples.}
    \label{fig:per_image_distributions}
\end{figure*}

The per-image distributions in Figure~\ref{fig:per_image_distributions} provide further insight into this behavior. Under in-dataset evaluation on CryoNuSeg, most methods exhibit relatively compact performance distributions. When transferred from CryoNuSeg to CPM, the distributions become substantially wider for several competing methods, indicating increased sensitivity to individual target images under domain shift. GDBA-Net maintains a comparatively high central tendency for Dice and Jaccard while avoiding the pronounced degradation observed for several baselines. The Precision and Specificity distributions further show that high background classification performance can coexist with substantially reduced segmentation overlap under cross-domain transfer. Overall, these results suggest that integrating pretrained semantic representations with frequency-derived structural information improves resistance to domain-specific appearance variations and enables more reliable segmentation of previously unseen histopathology data.

\subsection{Generalization vs target-domain adaptation}
\label{subsec:generalization_adaptation}

\begin{figure}[!t]
    \centering
    \includegraphics[width=0.95\columnwidth]{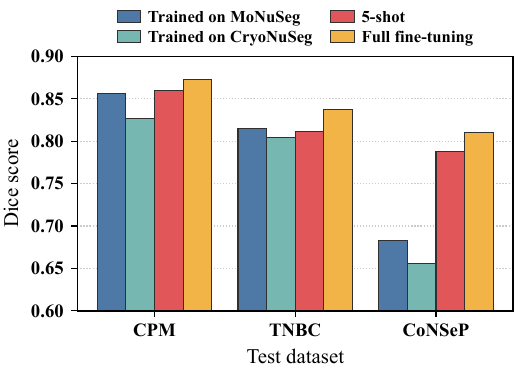}
    \caption{\textbf{Comparison of dice scores for GDBA across cross-dataset transfer, five-shot adaptation, and full target-domain fine-tuning on the CPM, TNBC, and CoNSeP datasets.} Cross-dataset models are trained on MoNuSeg or CryoNuSeg and evaluated directly on each unseen target dataset.}
    \label{fig:generalization_adaptation}
\end{figure}

Figure~\ref{fig:generalization_adaptation} compares the performance of GDBA-Net under direct cross-dataset transfer, 5-shot target-domain adaptation, and full target-domain training on CPM, TNBC, and CoNSeP. On CPM and TNBC, cross-dataset transfer remains close to few-shot and fully supervised performance, indicating that GDBA-Net transfers effectively when the source and target domains have comparable nuclear characteristics. In contrast, a larger performance gap is observed on CoNSeP, where direct transfer from MoNuSeg or CryoNuSeg results in lower performance. Introducing only five labeled target-domain images substantially narrows this gap towards the fully supervised performance. These results demonstrate that GDBA-Net can maintain strong generalization when domain shift is moderate, while limited target-domain supervision provides substantial benefits under more challenging domain shifts.

\subsection{Qualitative segmentation analysis}
\label{subsec:qualitative_analysis}

\begin{figure*}[!t]
    \centering
    \includegraphics[width=\textwidth]{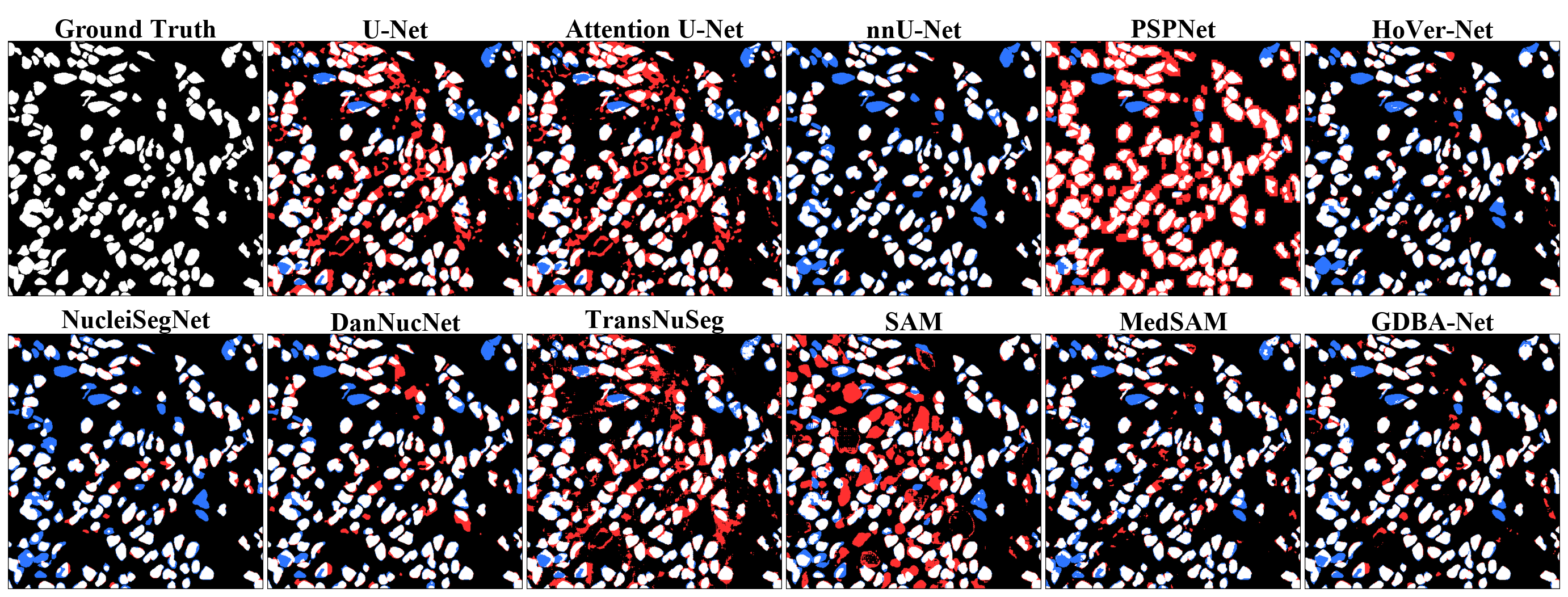}
    \caption{\textbf{Qualitative comparison of nuclei segmentation results on representative samples.} Ground-truth masks and overlap maps produced by the competing methods are shown for visual comparison. In the overlay, white denotes correctly segmented nuclei pixels, red denotes false positives predictions, and blue indicates false negatives predictions.}
    \label{fig:qualitative_segmentation}
\end{figure*}

Figure~\ref{fig:qualitative_segmentation} presents a qualitative comparison of segmentation results on representative samples. U-Net, Attention U-Net, PSPNet, and SAM exhibit false-positive regions, over-segmentation, or missed nuclei, particularly in densely populated regions. Although HoVer-Net and MedSAM yield stronger masks, some nuclei remain incomplete or missed. In comparison, GDBA-Net more closely follows the ground-truth masks, with improved boundary delineation and fewer false-positive and false-negative regions.

\subsection{Parameter Efficiency}
\label{subsec:parameter_efficiency}

\begin{figure}[!t]
    \centering
    \includegraphics[width=0.95\columnwidth]{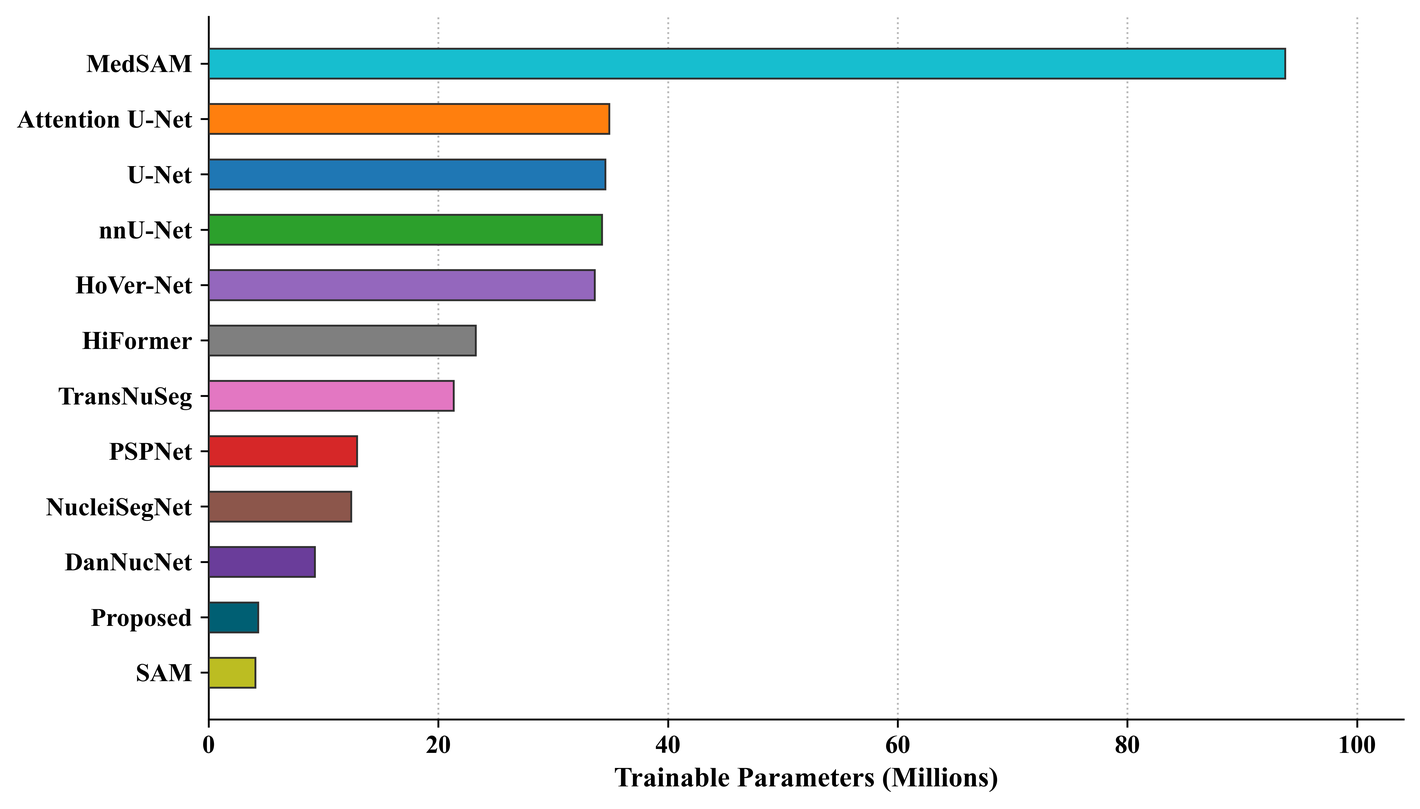}
    \caption{\textbf{Comparison of trainable parameter counts for GDBA and competing segmentation methods.}}
    \label{fig:parameter_efficiency}
\end{figure}

Despite leveraging the powerful SAM2 backbone, GDBA-Net restricts optimization to the lightweight GDBA modules while keeping the pretrained encoder frozen. As shown in Figure~\ref{fig:parameter_efficiency}, the proposed framework requires only 4.3 million trainable parameters, fewer than several competing models while maintaining strong segmentation performance. This favorable balance between trainable model capacity and segmentation accuracy highlights the parameter-efficient nature of the proposed adaptation strategy.

\subsection{Applicability to Additional Biomedical Segmentation Tasks}
\label{subsec:additional_tasks}

To examine the applicability of GDBA-Net beyond nuclei segmentation, we further evaluate the framework on two distinct biomedical segmentation tasks: retinal vessel segmentation using DRIVE~\cite{staal2004drive} and skin lesion segmentation using ISIC 2018 dataset~\cite{codella2019isic,tschandl2018ham10000}. DRIVE contains retinal fundus images with pixel-level vessel annotations, whereas ISIC 2018 contains dermoscopic images with corresponding skin lesion masks. These datasets provide additional evaluation of GDBA-Net across structures and imaging characteristics that differ substantially from histopathological nuclei.

As shown in Table~\ref{tab:additional_tasks}, GDBA-Net has better performance across both datasets and evaluation metrics. The strong performance across two structurally distinct segmentation tasks indicates that the benefits of integrating pretrained semantic representations with frequency-derived structural information are not limited to nuclei segmentation. This further supports GDBA-Net's adaptability across diverse biomedical segmentation tasks.

\begin{table}[!t]
\centering
\caption{Performance comparison of GDBA-Net and competing methods on ISIC 2018 and DRIVE for skin lesion and retinal vessel segmentation.}
\label{tab:additional_tasks}

\scriptsize
\setlength{\tabcolsep}{3.2pt}
\renewcommand{\arraystretch}{1.05}

\begin{tabular}{lcccc}
\hline
\multirow{2}{*}{\textbf{Models}} &
\multicolumn{2}{c}{\textbf{ISIC 2018}~\cite{codella2019isic,tschandl2018ham10000}} &
\multicolumn{2}{c}{\textbf{DRIVE}~\cite{staal2004drive}} \\
\cline{2-5}
&
\textbf{Dice} &
\textbf{Jaccard} &
\textbf{Dice} &
\textbf{Jaccard} \\
\hline
U-Net        & 0.858 & 0.776 & 0.816 & 0.701 \\
nnU-Net      & 0.855 & 0.766 & 0.780 & 0.640 \\
Attn U-Net   & 0.867 & 0.785 & 0.808 & 0.681 \\
NucleiSegNet & 0.881 & 0.808 & 0.650 & 0.482 \\
MedSAM       & 0.863 & 0.759 & 0.806 & 0.676 \\
SAM          & 0.903 & 0.827 & 0.741 & 0.588 \\
GDBA-Net     & \textbf{0.921} & \textbf{0.864} & \textbf{0.819} & \textbf{0.702} \\
\hline
\end{tabular}
\end{table}

\section{ABLATION STUDIES}
\label{sec:ablation}

\subsection{Contribution of GDBA components}
\label{subsec:gdba_contribution_ablation}

To systematically assess the contribution of the GDBA components, we evaluate multiple architectural configurations on the MoNuSeg and CryoNuSeg datasets (Table~\ref{tab:gdba_components}). Removing either the frequency or semantic branch reduces segmentation performance, demonstrating the complementary contribution of structural and semantic representations. Disabling the gating mechanism also degrades segmentation quality, confirming that adaptive weighting of the two feature streams is more effective than direct fusion. Replacing the shared projection with independent stage-specific projections increases the number of trainable parameters while reducing overlap performance. These findings demonstrate that the effectiveness of GDBA-Net stems from the coordinated contributions of its dual-branch representation, adaptive gating, and parameter-efficient shared-projection design.

\begin{table*}[!t]
\centering
\caption{Component-wise ablation analysis of the proposed GDBA framework on the MoNuSeg and CryoNuSeg datasets.}
\label{tab:gdba_components}

\scriptsize
\setlength{\tabcolsep}{3.2pt}
\renewcommand{\arraystretch}{1.05}

\begin{tabular}{ccccccccccccc}
\hline
\multirow{2}{*}{\begin{tabular}[c]{@{}c@{}}\textbf{Frequency}\\\textbf{branch}\end{tabular}} &
\multirow{2}{*}{\begin{tabular}[c]{@{}c@{}}\textbf{Semantic}\\\textbf{branch}\end{tabular}} &
\multirow{2}{*}{\begin{tabular}[c]{@{}c@{}}\textbf{Adaptive}\\\textbf{gate}\end{tabular}} &
\multirow{2}{*}{\begin{tabular}[c]{@{}c@{}}\textbf{Shared}\\\textbf{projections}\end{tabular}} &
\multirow{2}{*}{\begin{tabular}[c]{@{}c@{}}\textbf{Params}\\\textbf{(M)}\end{tabular}} &
\multicolumn{4}{c}{\textbf{MoNuSeg}} &
\multicolumn{4}{c}{\textbf{CryoNuSeg}} \\
\cline{6-13}
& & & & &
\textbf{Dice} & \textbf{Jacc.} & \textbf{Prec.} & \textbf{Spec.} &
\textbf{Dice} & \textbf{Jacc.} & \textbf{Prec.} & \textbf{Spec.} \\
\hline

$\times$ & $\checkmark$ & $\checkmark$ & $\checkmark$ &
4.11 & 0.8145 & 0.6893 & 0.8028 & 0.9332 &
0.8331 & 0.7156 & 0.8618 & 0.9369 \\

$\checkmark$ & $\times$ & $\checkmark$ & $\checkmark$ &
4.07 & 0.8087 & 0.6813 & 0.7954 & 0.9292 &
0.8360 & 0.7196 & 0.8248 & 0.9123 \\

$\checkmark$ & $\checkmark$ & $\times$ & $\checkmark$ &
4.28 & 0.8108 & 0.6841 & 0.7982 & 0.9306 &
0.8340 & 0.7173 & \textbf{0.8686} & \textbf{0.9396} \\

$\checkmark$ & $\checkmark$ & $\checkmark$ & $\times^{*}$ &
4.88 & 0.8048 & 0.6758 & 0.7907 & 0.9268 &
0.8402 & 0.7260 & 0.8568 & 0.9311 \\

$\checkmark$ & $\checkmark$ & $\checkmark$ & $\checkmark$ &
4.30 & \textbf{0.8408} & \textbf{0.7259} & \textbf{0.8388} & \textbf{0.9516} &
\textbf{0.8481} & \textbf{0.7381} & 0.8519 & 0.9286 \\

\hline
\end{tabular}

\vspace{1pt}
\raggedright
{\scriptsize $^{*}$ Independent stage-specific projections are used instead of the shared projection.}
\end{table*}

\subsection{Bottleneck Scaling Factor}
\label{subsec:bottleneck_scaling}

Table~\ref{tab:bottleneck_scaling} evaluates the trade-off between adapter capacity and parameter efficiency. Empirical evaluations indicate that $\alpha = 32$ strikes the optimal balance between expressiveness and regularized efficiency. Increasing capacity ($\alpha = 8$) increases the parameter count to 10.23M without improving segmentation, while stronger compression ($\alpha = 64$) slightly reduces both parameters and overlap performance.

\begin{table*}[!t]
\centering
\caption{Effect on bottleneck scaling factor $\alpha$ on model performance and parameter efficiency. Larger $\alpha$ corresponds to stronger dimensionality reduction.}
\label{tab:bottleneck_scaling}

\scriptsize
\setlength{\tabcolsep}{4pt}
\renewcommand{\arraystretch}{1.05}

\begin{tabular}{cccccccccc}
\hline
\multirow{2}{*}{\boldmath$\alpha$} &
\multirow{2}{*}{\textbf{Params (M)}} &
\multicolumn{4}{c}{\textbf{MoNuSeg}} &
\multicolumn{4}{c}{\textbf{CryoNuSeg}} \\
\cline{3-10}
& &
\textbf{Dice} & \textbf{Jacc.} & \textbf{Prec.} & \textbf{Spec.} &
\textbf{Dice} & \textbf{Jacc.} & \textbf{Prec.} & \textbf{Spec.} \\
\hline

8 &
10.23 &
0.8176 & 0.6941 & 0.8068 & 0.9347 &
0.8368 & 0.7213 & 0.8591 & 0.9332 \\

16 &
5.54 &
0.8189 & 0.6958 & 0.8087 & 0.9359 &
0.8352 & 0.7189 & \textbf{0.8618} & 0.9250 \\

32 &
4.30 &
\textbf{0.8408} & \textbf{0.7259} & \textbf{0.8388} & \textbf{0.9516} &
\textbf{0.8481} & \textbf{0.7381} & 0.8519 & \textbf{0.9286} \\

64 &
3.97 &
0.8194 & 0.6963 & 0.8096 & 0.9361 &
0.8372 & 0.7213 & 0.8475 & 0.9276 \\

\hline
\end{tabular}
\end{table*}
\section{DISCUSSION}
\label{sec:discussion}

The proposed framework improves nuclei segmentation by combining a frozen foundation model encoder with a lightweight domain adaptation mechanism that integrates pretrained semantic representations with frequency-derived structural information. While the gains under matched training and testing distributions were relatively modest, the advantage became considerably more evident under domain shift, where GDBA-Net achieved the highest Dice and Jaccard scores in seven of the eight evaluated source-target settings. This is particularly important in computational pathology, where variations in tissue type, preparation, staining, and acquisition can substantially alter image appearance and limit the generalizability of models optimized for a single dataset.

The cross-dataset results suggest that the complementary design of GDBA-Net is central to this robustness. SAM2 provides rich pretrained semantic representations, whereas the frequency branch emphasizes structural information associated with nuclear boundaries and local intensity transitions. Their integration allows the model to rely on information that is potentially less sensitive to source-specific appearance. This is especially evident in the strong transfer from CryoNuSeg to FFPE datasets despite their distinct preparation characteristics. Importantly, the CoNSeP-to-CPM setting, where HoVer-Net achieved higher overlap, also demonstrates that no representation is universally optimal and highlights the importance of evaluating generalization across multiple source-target combinations rather than isolated datasets.

The adaptation experiments further reveal a practical relationship between domain shift and annotation requirements. For CPM and TNBC, direct cross-dataset transfer of GDBA-Net remained close to target-domain-supervised performance, suggesting that limited or no additional target-domain annotation may be sufficient when cross-domain is already strong. For CoNSeP, where direct cross-domain transfer exhibited a larger performance gap, only five labeled target images recovered a substantial portion of the gap towards fully supervised performance. GDBA-Net therefore provides a practical continuum between direct cross-domain transfer, limited target-domain adaptation, and full target-domain training, which is valuable in histopathology where pixel-level annotation by experts is expensive and time-consuming.

The ablation studies reinforce that these gains originate from architectural design rather than increased trainable model capacity. Removing either feature branch or the gating mechanism consistently degraded overlap performance, confirming the benefit of complementary representations and their adaptive integration. Moreover, increasing the number of trainable parameters through independent projections or a wider bottleneck did not improve segmentation. With only approximately 4.3M trainable parameters, GDBA-Net demonstrates that foundation models can be effectively specialized for nuclei segmentation without full backbone fine-tuning. This combination of cross-domain robustness, adaptation with limited target-domain supervision, and parameter efficiency highlights the potential of the proposed design to specialize large pretrained models for fine-grained biomedical segmentation tasks.

Despite these findings, the present study is limited to H\&E nuclei segmentation using publicly available datasets. Although the additional experiments on retinal vessel and skin lesion segmentation provide initial evidence of applicability beyond histopathological nuclei, broader validation across imaging modalities, institutions, acquisition systems, and segmentation tasks are still needed to establish the generality of the approach. In addition, the current frequency extraction uses a fixed filtering strategy, and the effect of the frequency cutoff has not been systematically investigated. Learning or adapting the relevant frequency components may further improve performance across varying resolutions and noise characteristics. Future work will therefore explore adaptive frequency modeling, instance-level nuclear segmentation, and extension of the GDBA principle to other computational pathology and medical imaging tasks.
\section{CONCLUSION}
\label{sec:conclusion}

This study introduced GDBA-Net, a parameter-efficient adaptation framework built around the Gated Dual-Branch Adapter (GDBA), which integrates frequency-domain structural priors with pretrained semantic representations for robust nuclei segmentation. Through adaptive channel-wise gating and shared stage-specific projections, GDBA effectively incorporates complementary structural and semantic information into the hierarchical encoder. Evaluations across six heterogeneous histopathology datasets demonstrated accurate in-dataset segmentation, robust generalization to unseen domains, and effective adaptation using only five labeled target images. Additional evaluations on retinal vessel and skin lesion segmentation further demonstrated the applicability of the proposed adaptation strategy beyond nuclei segmentation. The ablation results further confirmed that these improvements arise from the coordinated design of the semantic and frequency branches, the gating mechanism, and the shared projection rather than increased trainable model capacity. 

More broadly, these findings demonstrate the value of combining task-relevant structural priors with transferable semantic representations. The strong cross-dataset performance of GDBA-Net suggests that their adaptive integration can improve robustness to domain-specific appearance variations while preserving fine structural information. GDBA-Net therefore provides a parameter- and annotation-efficient strategy for addressing two persistent challenges in computational pathology: domain variability and limited labeled data. Beyond nuclei segmentation, the results suggest that the proposed adaptation principle may provide a useful direction for specializing foundation models to fine-grained biomedical segmentation tasks where structural information and cross-domain robustness are important.
\section*{CRediT Authorship Contribution Statement}

Biplab Poudel (Conceptualization, Methodology, Data preparation, Formal analysis, Validation, Initial manuscript, Visualization), Betserai Albert Magarire (Data preparation, Formal analysis, Validation, Visualization), Anish Kumar Shah (Formal analysis, Validation), Jiacheng Xie (Formal analysis, Validation), Dong Xu (Conceptualization, Resources, Supervision, Funding acquisition). All authors contributed to the revision of the manuscript and have approved the final version.

\section*{Declaration of Competing Interests}

The authors declare that they have no known competing financial interests or personal relationships that could have appeared to influence the work reported in this paper.

\section*{Funding}

This work was partially supported by the National Institutes of Health (grants R35GM126985 and R01LM015252).

\section*{Code Availability}

The source code is available at: \url{https://github.com/biplabpoudel25/GDBA_Net}.

% References
% \section*{References}
\bibliographystyle{IEEEtran}
\bibliography{references}

\end{document}